\documentclass[sigconf]{acmart}
\AtBeginDocument{%
  \providecommand\BibTeX{{%
    \normalfont B\kern-0.5em{\scshape i\kern-0.25em b}\kern-0.8em\TeX}}}

\begin{document}
\title[Accurate and Clear Precipitation Nowcasting]{Accurate and Clear Precipitation Nowcasting with \\Consecutive Attention and Rain-map Discrimination}

\author{Ashesh}
\email{ashesh276@gmail.com}
\affiliation{%
  \institution{National Taiwan University}
  \city{Taipei}
  \country{Taiwan}
}

\author{Buo-Fu Chen}
\email{bfchen@ntu.edu.tw}
\affiliation{%
  \institution{National Taiwan University}
  \city{Taipei}
  \country{Taiwan}
}

\author{Treng-Shi Huang}
\email{tshuang@cwb.gov.tw}
\affiliation{%
  \institution{Central Weather Bureau}
  \city{Taipei}
  \country{Taiwan}
}

\author{Boyo Chen}
\email{boyochen722@gmail.com}
\affiliation{%
  \institution{National Taiwan University}
  \city{Taipei}
  \country{Taiwan}
}
\author{Chia-Tung Chang}
\email{r07229014@ntu.edu.tw}
\affiliation{%
  \institution{National Taiwan University}
  \city{Taipei}
  \country{Taiwan}
}
\author{Hsuan-Tien Lin}
\email{htlin@csie.ntu.edu.tw}
\affiliation{%
  \institution{National Taiwan University}
  \city{Taipei}
  \country{Taiwan}
}


\begin{abstract}
% OLD ABSTRACT
% Recently, there has been a lot of interest in Extreme Precipitation Nowcasting problem using deep learning techniques. Due to relative rarity of high rainfall events, weighted loss metrics have been formulated. In this work, we observe that such formulation adversely affects low rainfall predictions. Resulting blurry predictions also do not generate confidence in meteorologists which question their usability. We fix this practical issue by introducing a Discriminator which encourages the model to generate realistic rain maps without sacrificing predictive quality. Secondly, we introduce an attention based module which rescales the final prediction and achieves better performance on both high rainfall regions as well as lower rainfall regions. It is worthwhile to note that most works typically don't use actual rain data for target but instead work with transformed radar data thereby simplifying the problem. We work with real rain data and introduce a new dataset for Precipitation Nowcasting providing rain and radar data.

Precipitation nowcasting is an important task for weather forecasting. Many recent works aim to predict the high rainfall events more accurately with the help of deep learning techniques, but such events are relatively rare. The rarity is often addressed by formulations that re-weight the rare events. Somehow such a formulation carries a side effect of making ``blurry'' predictions in low rainfall regions and cannot convince meteorologists to trust its practical usability. We fix the trust issue by introducing a discriminator that encourages the prediction model to generate realistic rain-maps without sacrificing predictive accuracy. Furthermore, we extend the nowcasting time frame from one hour to three hours to further address the needs from meteorologists. The extension is based on  consecutive attentions across different hours. We propose a new deep learning model for precipitation nowcasting that includes both the discrimination and attention techniques. The model is examined on a newly-built benchmark dataset that contains both radar data and actual rain data. The benchmark, which will be publicly released, not only establishes the superiority of the proposed model, but also is expected to encourage future research on precipitation nowcasting.
\end{abstract}
%   <concept>
%       <concept_id>10010147.10010178.10010224</concept_id>
%       <concept_desc>Computing methodologies~Computer vision</concept_desc>
%       <concept_significance>300</concept_significance>
%       </concept>
%   <concept>
%       <concept_id>10010147.10010257.10010293.10010294</concept_id>
%       <concept_desc>Computing methodologies~Neural networks</concept_desc>
%       <concept_significance>300</concept_significance>
%       </concept>
%   <concept>
%       <concept_id>10010405.10010432.10010437</concept_id>
%       <concept_desc>Applied computing~Earth and atmospheric sciences</concept_desc>
%       <concept_significance>500</concept_significance>
%       </concept>
%   <concept>
%       <concept_id>10010147.10010257.10010258.10010259.10010264</concept_id>
%       <concept_desc>Computing methodologies~Supervised learning by regression</concept_desc>
%       <concept_significance>300</concept_significance>
%       </concept>
%  </ccs2012>

% \ccsdesc[300]{Computing methodologies~Computer vision}
% \ccsdesc[300]{Computing methodologies~Neural networks}
% \ccsdesc[500]{Applied computing~Earth and atmospheric sciences}
% \ccsdesc[300]{Computing methodologies~Supervised learning by regression}

\keywords{convolution neural networks, rainfall prediction, precipitation nowcasting, attention, discriminator, sequence models}


\maketitle

\section{Introduction}
Short-term (usually referred to as $<12$ hours) precipitation forecasting is one of the most important weather forecasting topics due to the ever-growing need for real-time, large-scale, and fine-grained precipitation nowcasting. Better short-term forecasts facilitate more efficient and safer daily lives; it helps provide road conditions, traffic jams, aviation weather report, and flood alert information to the society. Generally, for $6$ to $12$-hour forecasts, the numerical weather prediction (NWP) models driven by physics simulation provide superior and more stable predictions than conventional data-driven statistical techniques due to the refinement of model physics and computational schemes~\cite{kain2010assessing,sun2014use}. For $0$ to $1$-hour quantitative precipitation nowcasting (QPN), on the other hand, radar echo extrapolation remains a powerful and highly relevant method~\cite{dixon1993titan,germann2002scale,germann2004scale,chung2020improving} because of the high temporal and spatial resolutions of radar maps whenever they are available. However, the major drawback of these extrapolation-based QPN techniques includes the difficulty to capture the growth and decay of storms, uncertainty in converting radar reflectivity to actual rainfall amount, and limitation of anticipating storm motion at the larger lead time (i.e., the $2^{nd}$ and $3^{rd}$ hours). 

The deep learning community has recently shown great interest in the QPN problem with several recent works~\cite{shi_convolutional_2015,shi_deep_2017,tran2019computer,franch_precipitation_2020}. Notably, Shi et al.~\cite{shi_convolutional_2015} tackled this problem by modeling it as a spatiotemporal sequence forecasting problem (i.e., predicting the animation of radar echo), introducing the encoding-forecasting structure of ConvLSTM. Another recent work~\cite{shi_deep_2017} replaced ConvLSTM with a novel recurrent module, TrajGRU, while using a similar encoding-forecasting structure. The TrajGRU module was claimed to learn location-variant structure.
Although these studies have demonstrated a great potential to apply deep learning to QPN, two critical issues need to be addressed.
First, the deep learning model should predict heavy rainfall and simultaneously keep good performance in drizzle areas. If the model demonstrates good performance on heavy rainfalls but outputs an unrealistically large rain area, the forecaster may get confused (why is it raining everywhere?) and lose trust in the deep-learning-driven QPN system.
In addition, most of the previous QPN models~\cite{shi_convolutional_2015,shi_deep_2017,tran2019computer,franch_precipitation_2020} provide only a one-hour prediction that is not long enough for forecasters. Extending the prediction period from one hour to three hours can be a unique selling proposition of deep learning models compared to the conventional radar echo extrapolation method.

In this work, we tackle both issues above by designing a novel deep learning model. We first address the issue of producing predictions that humans can trust in both low rainfall regions and high rainfall ones. We observe that it is easy for a human to distinguish between a real rain map and the predicted (generated) rain map, where visual blurriness appears to be the primary distinguishing factor. We thus design a discriminator that learns to distinguish between real and generated rain maps. The discriminator loss nudges the deep model to generate more realistically looking rain maps without compromising model performance. We observe both qualitatively and quantitatively that this indeed leads to generated rain maps with less blurriness to win human trust. 

Then, we address the issue of making accurate $1$ to $3$-hour predictions by proposing an attention-based model. The attention mechanism creates focus regions on the rain maps that can be carried from the earlier hours to later ones, and the focus regions are used to rescale the predicted rainfall values. We observe that this leads to consistent improved performance on both low and high rainfall regions. 

Combining the discriminator and the attention mechnism results in a novel deep learning model, which, to the best of our knowledge, is the first deep learning model that tackles the extreme precipitation nowcasting problem using from both radar and real rain data. The dataset that contains both part of the data is publicly released to encourage more research in this direction. Experiments on the dataset justify the validity of our designs on using the discriminator and attention mechanism, and demonstrate the superiority of the proposed deep learning model.

The paper is organized as follows. Section 2 describes the related works regarding QPN models and deep learning components related to our proposed model architecture. Section 3 first describes the radar and rain rate data used in this study and later describes the different components of our model which involves discriminator-based architecture, attention module and different loss components. Section 4 evaluates the performance of our model using statistical analysis and a case study. At the end, our final remarks and future work is stated in Section 5.

\section{Related Work}
As mentioned in the first section, Shi et al.~\cite{shi_convolutional_2015} introduced an encoding-forecasting structure to model QPN as a spatiotemporal sequence forecasting problem. Their following work~\cite{shi_deep_2017} replaced ConvLSTM with a TrajGRU module to learn the location-variant features in the radar map. To ensure better prediction for less frequent but heavy rainfall, they introduced a weighted loss, which gave more weightage to loss happening in heavy rainfall regions. 
However, these studies did not use rainfall observation as the target. Their models only predicted the radar echo in the future and coverted the predicted radar echo to rainfall using simplified relationships. This is a serious shortcoming in QPN since it questions its real world applicability.

Additionally, the Shi et al. model \cite{shi_deep_2017} suffers from under predictions for high rainfall regions despite having the weighted loss. Another issue is that the prediction contains large number of pixels with small amount of rainfall which gives a hazy appearance. This is a natural consequence of giving low weightage to low rainfall regions. 
They also use a mask to ignore pixels where rainfall is below a threshold. While this mask-based weighted loss formulation is very useful for training the highly skewed rain data, the haziness in prediction is not desirable. The haziness in prediction makes meteorologists skeptical about using the deep learning model since the conventional QPN models used by meteorologists do not suffer from such artifacts. 

Subsequently, Tran and Song \cite{tran2019computer} tried to improve the ConvLSTM and TrajGRU QPN models by modifying the loss function design. They employed some visual image quality assessment techniques, including structural similarity (SSIM) and multi-scale SSIM, to train the models and significantly reduce the blurry image issue. Furthermore, Franch et al \cite{franch_precipitation_2020} proposed a solution based on model stacking to improve deep learning QPN prediction skills. They used a convolutional neural network to combine an ensemble of deep learning models focusing on various rain regimes, doubling the prediction skills. However, both of these studies only provide one-hour predictions, while the meteorological community genuinely needs deep learning QPN models with a longer prediction time.
  
To manage blurriness of the prediction and improve the QPN quality in high rainfall regions, this study touches upon two important techniques from deep learning, namely adverserial learning greatly popularized by generated adverserial networks (GAN) and attention. While GANs~\cite{goodfellow_generative_2014} are primarily used for image generation, their discriminator based adverserial learning approach has been used in a stand alone fashion to make the distribution of generated images similar to a desired distribution ~\cite{jakab_self-supervised_2020}. Thus the adverserial training leads to clarity in generated images when clarity is part of the desired distribution~\cite{vondrick_generating_2016,kwon_predicting_2019}. 
Attention-based approaches, while being initially developed for natural language processing tasks, have found decent popularity in the computer vision community~\cite{li_object_2020,Woo_2018_ECCV}. Their ability to focus differently for different pixels have been shown to do well in cases where features of interest spans small number of pixels and their occurrences are rare~\cite{bai_benchmarking_2020,lim_small_2019,zhang_multiresolution_2020}.
\begin{table}
  \begin{tabular}{ccccccccc}
    \toprule
    RR & 0-1 &   1-3 & 3-5 & 5-10 & 10-20 &20-30 & 30-40 & 40- \\
    \midrule
    \%        & 95.92 & 2.18& 0.71 & 0.67 & 0.37 & 0.09 & 0.03&0.03\\
    \bottomrule
  \end{tabular}
  \caption{Distribution of rainfall in QPESUMS dataset. RR is rain rate in mm/hr}
   \label{tab:rain_dist}
\end{table}
\section{Datassets: radar and rainfall datasets from Central Weather Bureau, Taiwan }
Two datasets are used in this work--- rain rate and radar reflectivity in Taiwan region. Both of the datasets are produced by the Quantitative Precipitation Estimation and Segregation Using Multiple Sensor (QPESUMS) system \cite{C.P-L_QPESUMS_2020} from Central Weather Bureau (CWB), Taiwan. QPESUMS provides high-resolution and rapid-updated rainfall data and radar reflectivities based on the observation from different radar site.

\paragraph{\textbf{Rain rate}}
The rainfall data is a 2D map of shape $561*441$ containing rain rate. The unit is $mm/hr$.

\paragraph{\textbf{Radar reflectivity}}
The 3D radar data possesses a horizontal size of $561*441$ and 21 levels in height. We take the maximum value of radar reflectivity (unit: $dBZ$) over the 21 channels to get 2D data which we work with.

Both rain and radar data have a time resolution of 10 minutes. We have this data from January 2015 till December 2018 comprising of 203K frames. We divide the data into train, test and validation by their timestamp:
\begin{enumerate}
    \item Training set: 2015-2017.
    \item Validation set: First 15 days of each month in 2018.
    \item Test set: Last 15 days of each month in 2018.
\end{enumerate}
This validation-test allocation ensures that both our test and validation data cover all seasons. Distribution of rainfall can be seen in ~Table\ref{tab:rain_dist}.

\paragraph{\textbf{Benchmark QPESUMS extrapolation model}}
The QPESUMS system also provides a 1 hour rain rate prediction which serves as our benchmark. The rain rate product is made by a radar extrapolation method, similar to optical flow \cite{QPF_2003, QPF_2010}. This extrapolation technique first analyzes the moving vector of each convection cell from the 2D reflectivity mosaic and then add the vectors onto the previous rainfall map. Because this technique doesn't take the evolution of convective cells into consideration, the growth and decline of weather systems will not be reflected in this benchmark. Thus, the effective forecast time is only up to 1 hour. This QPN product is currently being used operationally  by weather forecasters in the CWB, Taiwan.

\section{Proposed Approach}
In this section, we explain our proposed approach. First we formally introduce the problem setup. Next, we go on to describe the different components of our model. Thereafter, we describe our different loss components. Our model predicts $T=3$ hours into the future. For that we predict $T$ frames, one for each hour. As input, we feed last 60 minutes of rain and radar data to the model.

\subsection{Problem Setup}
Let $Rn_t$ and $Rd_t$ denote the 2D rain rate data and 2D radar data at time $t$ respectively where $t$ is measured in minutes. Let 
$Y_t=1/6 * \sum_{i=0}^5Rn_{t +10*i}$. 

Given $\{Rn_{t-10}$, $Rn_{t-20}$ .. $Rn_{t-10*N} \} $ and $\{Rd_{t-10}$,.. $Rd_{t-10*N} \} $, task is to predict $[Y_t,Y_{t+60},Y_{t+120}]$. In our setting, $N=6$. Put simply, given one hour rain and radar data, we want to predict average hourly rain-rate for next three hours.

\begin{figure}[t]
  \centering
  \includegraphics[width=\linewidth]{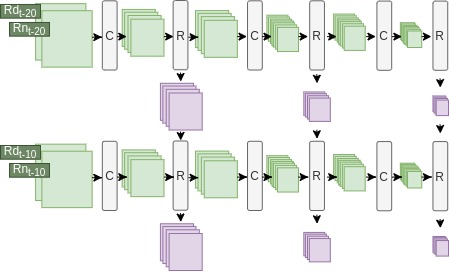}
  \caption{Data flow in Encoder. (R: RNN layer, C: Convolution layer with Leaky ReLU activation). Figure is shown for sequence length $N=2$}
  \label{fig:encoder}
\end{figure}

\begin{figure}[h]
  \centering
  \includegraphics[width=\linewidth]{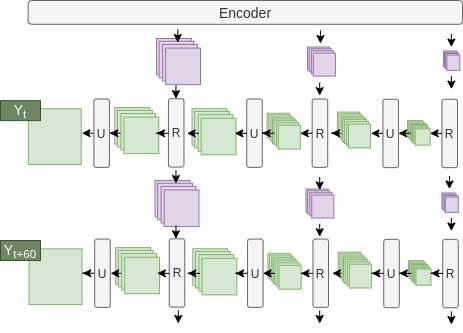}
  \caption{Data flow in Decoder. (R: RNN layer, U: Transposed convolution layer). Figure is showing 2 hour prediction. }
  \label{fig:decoder}
\end{figure}
\begin{figure}[t]
  \centering
  \includegraphics[width=\linewidth]{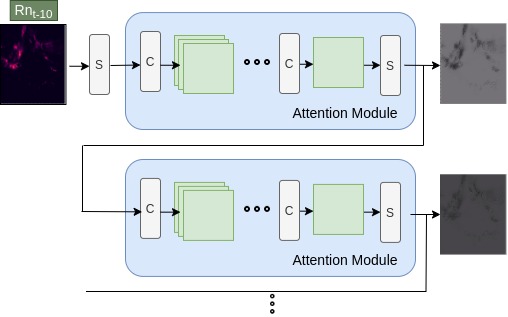}
  \caption{Attention Module (S: Sigmoid layer, C: Convolution layer)}
 \label{fig:attention}
\end{figure}

\begin{figure}[]
  \centering
  \includegraphics[width=\linewidth]{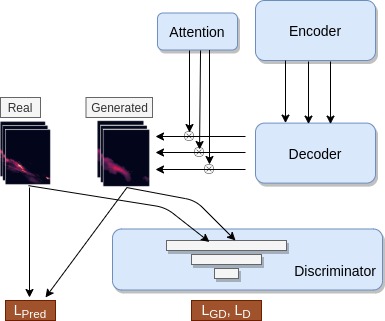}
  \caption{Overall Model}
 \label{fig:overall}

\end{figure}

\subsection{Model Components}
We next describe the three modules which comprise our model. The overall architecture is shown in Figure~\ref{fig:overall}.

\subsubsection{\textbf{Prediction Module}}
Our prediction module is inspired from the one used in~\cite{shi_deep_2017} and is an encoder-forecaster based architecture. We use ConvGRU (Convolution GRU) instead of their proposed TrajGRU for the RNN module as the latter did not give any extra benefit. We also introduce a spatial attention component described in the next subsection.  Encoder is composed of 3 layers of RNN module with one 2D convolution layer between every consecutive RNN module pair. Purpose of the 2D CNN module is to downsample the spatial dimension thereby allowing feature extraction at multiple scales. The forecaster has a similar structure with 3 RNN modules with two 2D transposed convolution modules sandwiched between them. In forecaster, the transposed convolution performs upsampling. One can find the schematic for Encoder and Decoder in Figure~\ref{fig:encoder} and Figure~\ref{fig:decoder}.

\subsubsection{\textbf{Attention Module}}
The attention module, shown in Figure~\ref{fig:attention} pixel-wise scales the prediction obtained from the Predicion module. It predicts an attention map for next hour rain-rate and takes as input the attention map for previous hour. For predicting the first hour attention map, sigmoid applied on the latest available rain map is taken as input. The module is composed of 2D convolution layers and uses leaky relu as activation. In this work, we refer to the presence of Attention module by token 'Atn'.

\subsubsection{\textbf{Discriminator Module}}
Together with the Prediction module, we also simultaneously train a Discriminator. It learns to discriminate between real rain maps and generated rain maps. The loss from the discriminator nudges the Prediction module towards generating realistic rain maps. This leads to a much clearer and therefore more informative prediction even with weak input signals as is the case for second and third hour prediction. Our Discriminator has a simple linear structure and is composed of Dense layers.

\subsection{Loss functions}

\subsubsection{\textbf{Prediction Loss}}
Following~\cite{shi_deep_2017}, we adopt a weighted loss scheme. Let $Y_t$ represent the actual $t^{th}$ one hour averaged rain rate 2D map sequence and $\hat{Y_t}$ represent the predicted version. We use a weighted MAE loss defined as follows:
\begin{equation}
    L_{Pred} = 1/(3*540*420)\sum_{t}^3 \sum_j^{540}\sum_k^{420} W(Y_t[j,k];0.5)* |Y_t[j,k] - \hat{Y}_t[j,k]|
    \label{eq:pred_loss}
\end{equation}
Here, Weight $W$ is defined as follows.
\begin{equation}
    W(x; Th) = \begin{cases}
    0 & x< Th\\
    1 & Th<=x<2\\
    2 & 2<=x<5\\
    5 & 5<=x<10\\
    10 & 10<=x < 30\\
    30 & 30<=x\\
    \end{cases}
\label{eq:loss_weight}
\end{equation}
We refer to this loss component as 'WMAE' (Weighted MAE). For comparison, we also use an MSE based version of it where MAE is replaced with MSE. We call it 'WMSE'. In both cases, unless explicitly stated, the threshold $Th$ is set to $0.5$

\subsubsection{\textbf{Discriminator Loss}}
From equations \ref{eq:pred_loss} and \ref{eq:loss_weight}, one can infer that model simply does not care about prediction for pixels where the target rain is less than 0.5mm/hr. As seen in the Figure~\ref{fig:Fig_case3h} (WMAE column), one can observe that this causes blurry prediction for second and third hour. We use a Discriminator to encourage the model to generate rain maps which look like the real ones. Discriminator itself is trained using the cross-entropy formulation as described below.
\begin{equation}
  L_D = -1*\sum_{t} \log(D(Y_t)) + \log(1-D(\hat{Y}_t))
\end{equation}
The discriminator loss formulation is shown below. Note that this loss component does not change the Discriminator's weights. It changes $\hat{Y_t}$ by updating Predictive module's weights so that $D(\hat{Y}_t)$ gets closer to 1.
\begin{equation}
    L_{GD} =-1*\sum_{t}\log(D(\hat{Y}_t))
\end{equation}
We refer to the Discriminator module and the two associated loss components $L_D, L_{GD}$ with 'Adv' token. 
Final loss expression for the Predictive module is the weighted sum of discriminator loss and the prediction loss.
\begin{equation}
    L_{G}= (1-w_{Adv})*L_{Pred} + w_{Adv}*L_{GD}
\end{equation}

\subsubsection{\textbf{Balanced Loss}}
This loss component was developed for comparison of Discriminator based approach and we do not use this in our final model. The motivation here is to account for pixels which are not considered in equations \ref{eq:pred_loss} and \ref{eq:loss_weight}--- pixels with rainfall values less than 0.5mm/hr. Let $Mask_{t}$ be a \{0,1\} valued tensor of same shape as $Y_t$ and $Mask_t[i,j]=1$ if and only if $Y_t[i,j]<0.5$. Balanced loss component, $L_{Bal}$ is then defined as 
\[L_{Bal} = 1/(540*420*3)\sum_{t}^3 \sum_j^{540}\sum_k^{420} Mask_t[j,k]* |Y_t[j,k] - \hat{Y}_t[j,k]|\].
The expression of loss for this model (GRU+WMAE+Bal) is  \((1-w_{Bal})*L_{Pred} + w_{Bal}*L_{Bal}\).

\subsection{Data Preprocessing}
For memory and file system management, we converted the rain and radar values to integer values. We center cropped the rain and radar data to size $540*420$. We normalized rain and radar data by dividing them with their respective 95th quantile value.

\subsection{Implementation Details}
The code is written using $Pytorch Lightning==1.0.2$ and $Pytorch==1.6.0$. We used Adam as optimizer with 0.0001 as the learning  rate. For all experiments, batch size is set to 16 and we train for a maximum of 15 epochs. We pick the best performing model on validation data using $L_{Pred}$ metric. 
Our Discriminator model is composed of 3 dense layers of sizes 128, 128 and 1. LeakyRelu is used for non-linearity except for the last layer where a sigmoid is used. For attention module, we use 5 Convolution layers whose output channel counts are [16,32,32,32,1]. Kernel size of 5 with padding of 2 is used. LeakyRelu is used for non-linearity after every convolution layer except the last one. For models having Discriminator, $w_{Adv}=0.05$ and for model using Balanced loss, $w_{Bal}=0.01$. They are obtained by looking at performance of validation set on WMAE metric.
\begin{table*}
  \begin{tabular}{c| cccccc| cccccc}
    \toprule
    Model & \multicolumn{6}{c}{2018}& \multicolumn{6}{c}{August 2018}\\
    & \multicolumn{3}{c}{WMAE(Th=0.5)} & \multicolumn{3}{c}{WMAE(Th=0.0)} &\multicolumn{3}{c}{WMAE(Th=0.5)} & \multicolumn{3}{c}{WMAE(Th=0.0)} \\
    & H0&H1&H2& H0&H1&H2& H0&H1&H2& H0&H1&H2\\
    \midrule
    % Last 20min &0.81&1.21&1.3&0.83&1.27&1.39&4.71&7.2&7.82&4.77&7.4&8.12\\
    Last 10min &0.75&1.2&1.3&0.76&1.27&1.38&4.36&7.16&7.82&4.4&7.36&8.1 \\
    QPESUMS    &0.96&1.19&1.32&1.06&1.3&1.49&   5.55&6.89&7.9&5.81&7.21&8.41\\
    TrajGRU + WMSE+WMAE~\cite{shi_deep_2017} & 0.57&0.89&0.99&2.97&5.85&7.23&     3.18&5.09&5.64&5.21&9.4&11.07 \\
    TrajGRU + WMAE&0.48&0.84&0.95&1.99&3.7&4.71&   2.65&4.84&5.56&3.91&7.5&9.19 \\
    \midrule
    GRU + WMAE & 0.49&0.84&0.95&1.78&3.51&4.57&   2.69&4.84&5.55&3.78&7.28&8.96 \\
    GRU + WMAE+Bal &0.49&0.84&0.95&0.66&1.4&1.91&   2.7&4.8&5.52&3.13&6.31&7.99 \\
    GRU + WMAE+Adv &0.49&0.85&0.96&0.58&1.26&1.85&   2.72&4.83&5.52&2.96&5.98&7.68\\
    % GRU + WMAE+Atn &0.48&0.84&0.95&2.17&4.27&5.29&    2.76&4.98&5.71&4.02&7.8&9.38\\
    GRU + WMAE+Adv+Atn &\textbf{0.48}&\textbf{0.84}&0.96&\textbf{0.55}&\textbf{1.19}&1.64&  
    \textbf{2.64}&4.81&\textbf{5.5}&\textbf{2.85}&\textbf{5.84}&\textbf{7.38}\\
    \bottomrule
  \end{tabular}
  \caption{Performance on WMAE metric for first (H0), second (H1) and third (H2) hour prediction.}
   \label{tab:wmae}

\end{table*}

\begin{figure*}[t]
  \centering
  \includegraphics[width=0.92\textwidth]{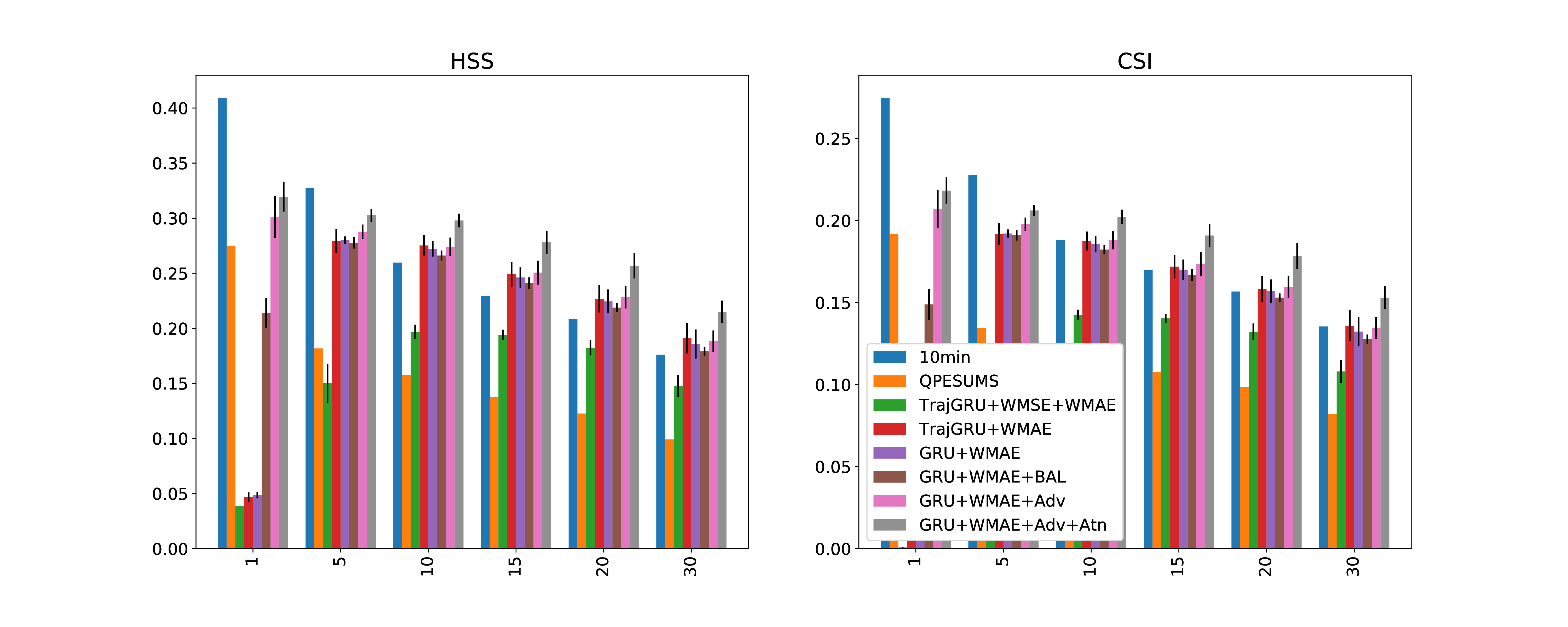}
  \caption{HSS and CSI metrics on 2018 data for first hour prediction}
  \label{fig:hss_csi_2018}
\end{figure*}

\begin{figure*}[]
  \centering
  \includegraphics[width=0.92\textwidth]{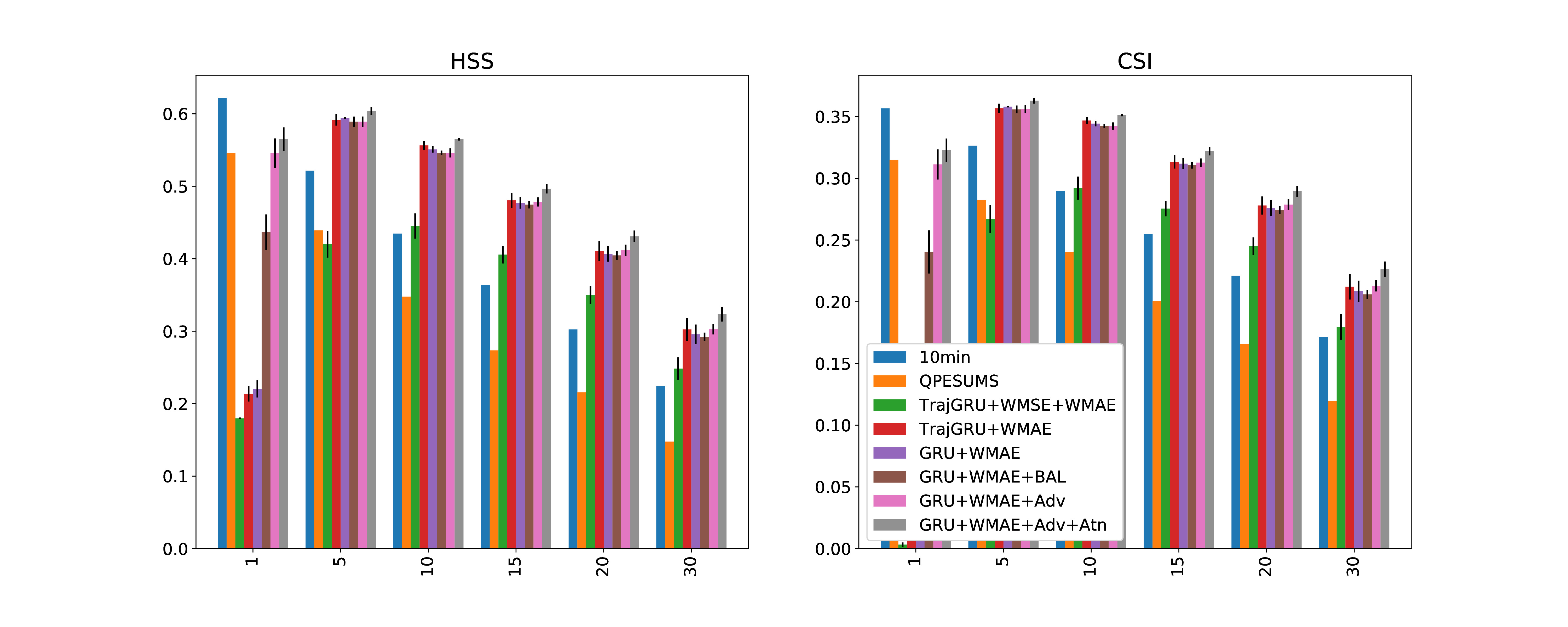}
  \caption{HSS and CSI metrics on August 2018 data for first hour prediction}
  \label{fig:hss_csi_aug}
\end{figure*}

\section{Evaluation}

\subsection{Evaluation Criteria}
We use multiple metrics to evaluate the performance of our model quantitatively and show a case study for qualitative evaluation. Firstly, we use the ${WMAE}$ with $Th=0.5$. As mentioned before, one issue with this metric is that it hides the poor performance on low rain regions. We therefore also evaluate $WMAE$ with $Th=0$. Since meteorologist are interested in assessing how the model performs for different amount of rain, we binarize our prediction with different thresholds and compute CSI and Heidke Skill Score (HSS)~\cite{hogan_equitability_2010}. For computing these metrices for a threshold, both prediction and target are binarized using the threshold. We then compute True positive (TP), False positive (FP), True Negative (TN) and False Negative (FN). CSI and HSS are computed as \(TP/(TP+FN+FP)\) and \((TP*TN -FP*FN)/((TP+FN)(TN+FN) + (TP+FP)(TN+FP))\) respectively. For both metrics, the higher the value, better is the model. Since our focus is on heavy rainfall and to that end we introduced the attention module, we evaluate the performance on two date ranges--- on whole of 2018 and on August 2018, the most rainy month of 2018.  We also plot a performance diagram depicting model's performance on different rainfall levels. 

For benchmarks, we train the model developed in~\cite{shi_deep_2017}. We refer to it by 'TrajGRU+WMAE+WMSE' since it uses TrajGRU~\cite{shi_deep_2017} layer for RNN module and uses average of weighted MAE and MSE as the loss. With~\cite{shi_deep_2017}, we observe that the WMSE loss component was causing a significant downgradation in performance. So, we train another variant of it where we just keep WMAE as the loss function, which we refer to by 'TrajGRU+WMAE'.  As discussed previously, we obtain another benchmark from CWB, Taiwan and refer to it as QPESUMS. The benchmark has the prediction for first hour only. For evaluating on second and third hour, we use this first hour prediction as the prediction for all 3 hours. Finally, as yet another benchmark, we use latest available rain map as  prediction which we refer by "Last 10min'.

\subsection{Overall statistics}
Performance on first hour prediction in terms of HSS and CSI metrics is given for 2018 in Figure~\ref{fig:hss_csi_2018} and for August 2018 in Figure~\ref{fig:hss_csi_aug}. Corresponding tables (Table~\ref{tab:hss_csi_2018},~\ref{tab:hss_csi_2018Aug}) are given in supplemental data. For thresholds greater than 5, we observe best performance in our Attention based model in both cases. The benefit is quite significant if one compares it with~\cite{shi_deep_2017} and with QPESUMS, a benchmark used by professional meteorologists.  The evidence of our Attention based model outperforming other deep learning models on all rain-rate thresholds (both high-rain and low-rain) can be seen in both these figures. The benefit of Attention module exclusively can be observed by comparing 'GRU+WMAE+Adv+Atn' vs 'GRU+WMAE+Adv'. Note that everything including all hyperparameters are identical for these two models except the presence/absence of attention module.

Our claim that the discriminator based approach gets clearer predictions can be inferred by observing its ('GRU+WMAE+Adv' model) superior performance on threshold of 1 with respect to models not having the discriminator ('Adv' token). It is worth noting that it gets similar if not better performance on higher thresholds in both Figures~\ref{fig:hss_csi_2018} and \ref{fig:hss_csi_aug}. We developed another model 'GRU+WMAE+Bal' for obtaining clarity in prediction whose details are given in Section 4.3. We see that the discriminator based model outperforms this model as well on threshold of 1.  All this can also be observed from performance on WMAE reported in Table~\ref{tab:wmae}. Here performance is shown for $Th=0.5$ and $Th=0$. Note that $Th=0$ is presented to quantify clarity in predictions since in this case, loss on pixels with hourly rain-rate < 0.5mm/hr are also included. By inspecting $Th=0$, one can observe that using discriminator (models with 'Adv') improves clarity.

On HSS and CSI metrics, we observe a decent performance of 'Last 10min' benchmark where we use the latest available rain map as prediction. We note that this especially works well for non-moving rainfall occurrences and low thresholds. It is so because with low thresholds, the model need not account for change in intensity of the rainfall with time. On the other hand, with WMAE metric shown in Table~\ref{tab:wmae} where the change in rainfall intensity is directly penalized, we naturally see worse performance of Last 10min benchmark, especially with $WMAE(Th=0.5)$. With $WMAE(Th=0.0)$, the deep learning based models are at a disadvantage since the uncertainty gets manifested in blurry predictions which mostly hampers performance on significant number of non-rainy pixels. 

\subsubsection{\textbf{A Blended Model}}
\begin{figure}
  \centering
  \includegraphics[width=\linewidth]{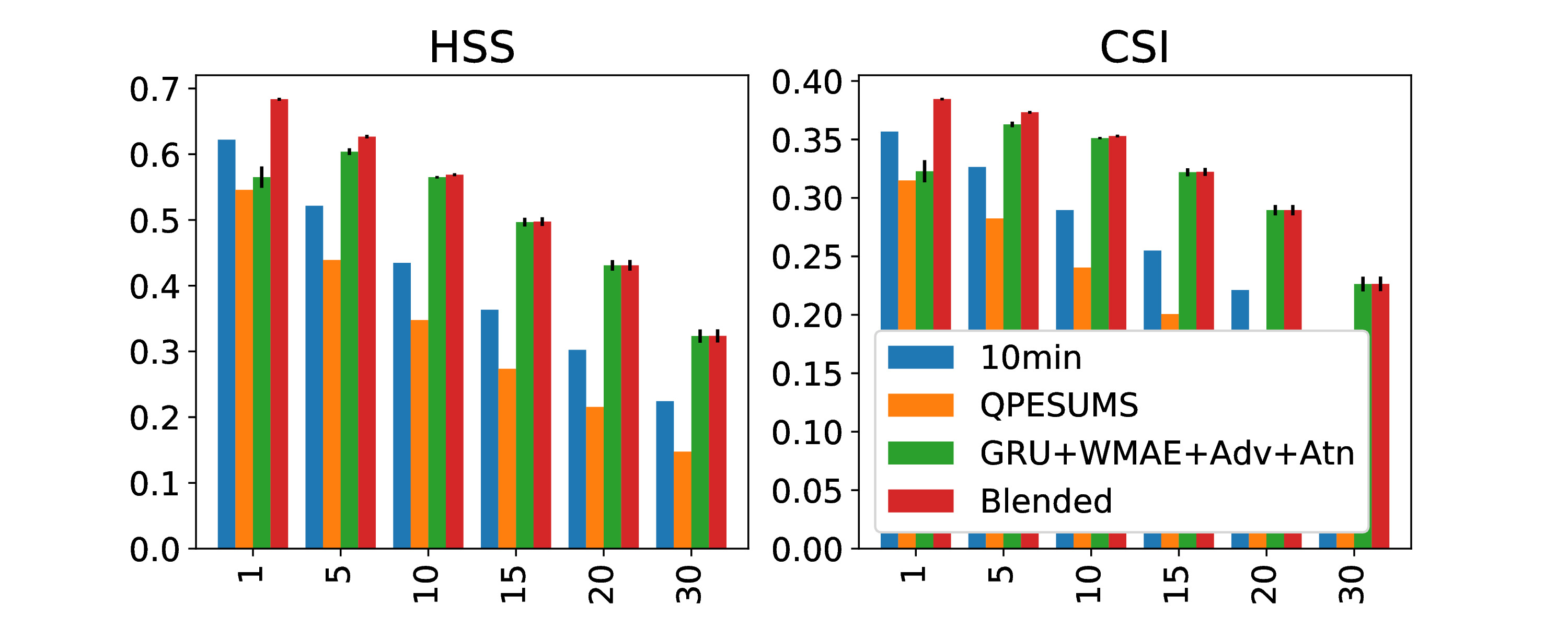}
  \caption{Blended model performance on August 2018 data for first hour prediction}
  \label{fig:hss_csi_aug_blended}
\end{figure}

Beyond clarity, which our model achieves (Figure~\ref{fig:Fig_case}), meteorologists are not concerned with very low rain-rate thresholds. Nonetheless, we argue that if needed, it is relatively easy to get even better performance on lower thresholds and to that end we showcase a blended version of our model. We first create a classification model which predicts whether or not a pixel will have hourly rain-rate exceeding 0.5mm/hr. Our final prediction for the blended version is the weighted average of prediction of GRU+WMAE+Adv+Atn model and last 10min rain-rate. Weights are computed using classifier's prediction. As seen in Figure~\ref{fig:hss_csi_aug_blended}, we see that the blended model outperforms the Last 10min benchmark on lower thresholds and achieves similar performance to our best model on higher thresholds. We don't prefer the blending, since similar to Last 10min benchmark, it is slightly inferior on WMAE metric.  Please refer to supplemental for more details.

\begin{figure}[t]
  \centering
  \includegraphics[width=\linewidth]{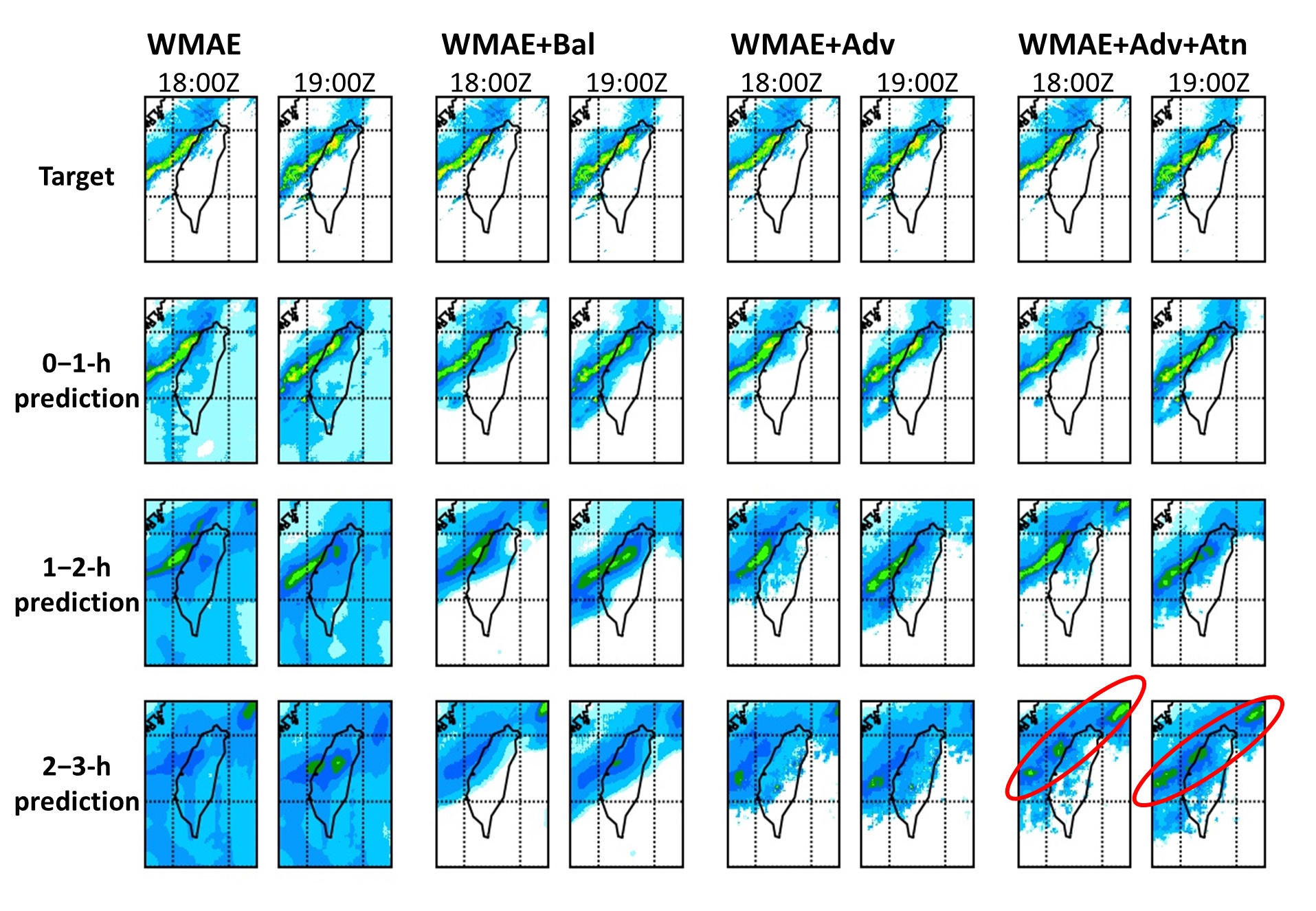}
  \caption{The observational rainfall (target) and the predictions that are conducted one, two, and three hours ago for four of our deep learning QPN models.}
  \label{fig:Fig_case3h}
\end{figure}

\begin{figure}[t]
  \centering
  \includegraphics[width=\linewidth]{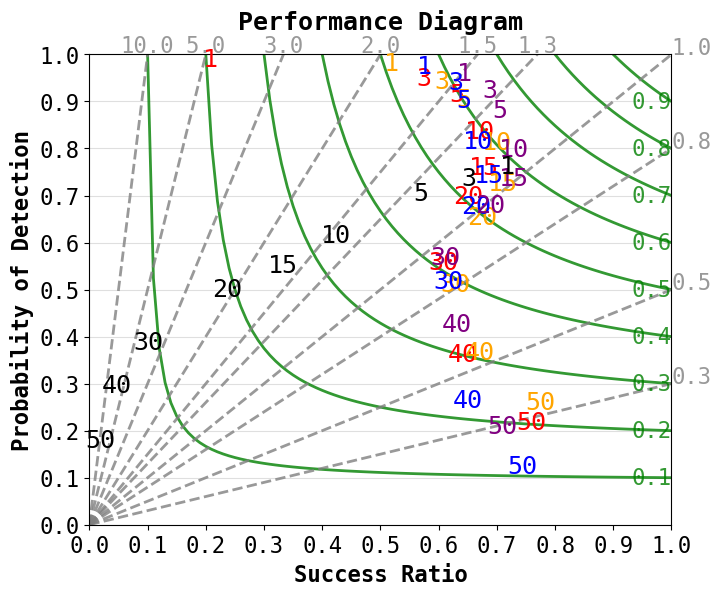}
  \caption{Performance diagram of +1 h prediction for the benchmark QPESUMS radar echo extrapolation (black text) and four of our deep learning QPN models for the front event. Red, gold, blue, and purple texts indicates the scores of different thresholds for GRU+WMAE, GRU+WMAE+Bal, GRU+WMAE+Adv, and GRU+WMAE+Adv+Atn model, respectively. }
 \label{fig:PD_1hr}
\end{figure}

\subsection{A case study}
 
\begin{figure*}[t]
  \centering
  \includegraphics[width=\textwidth]{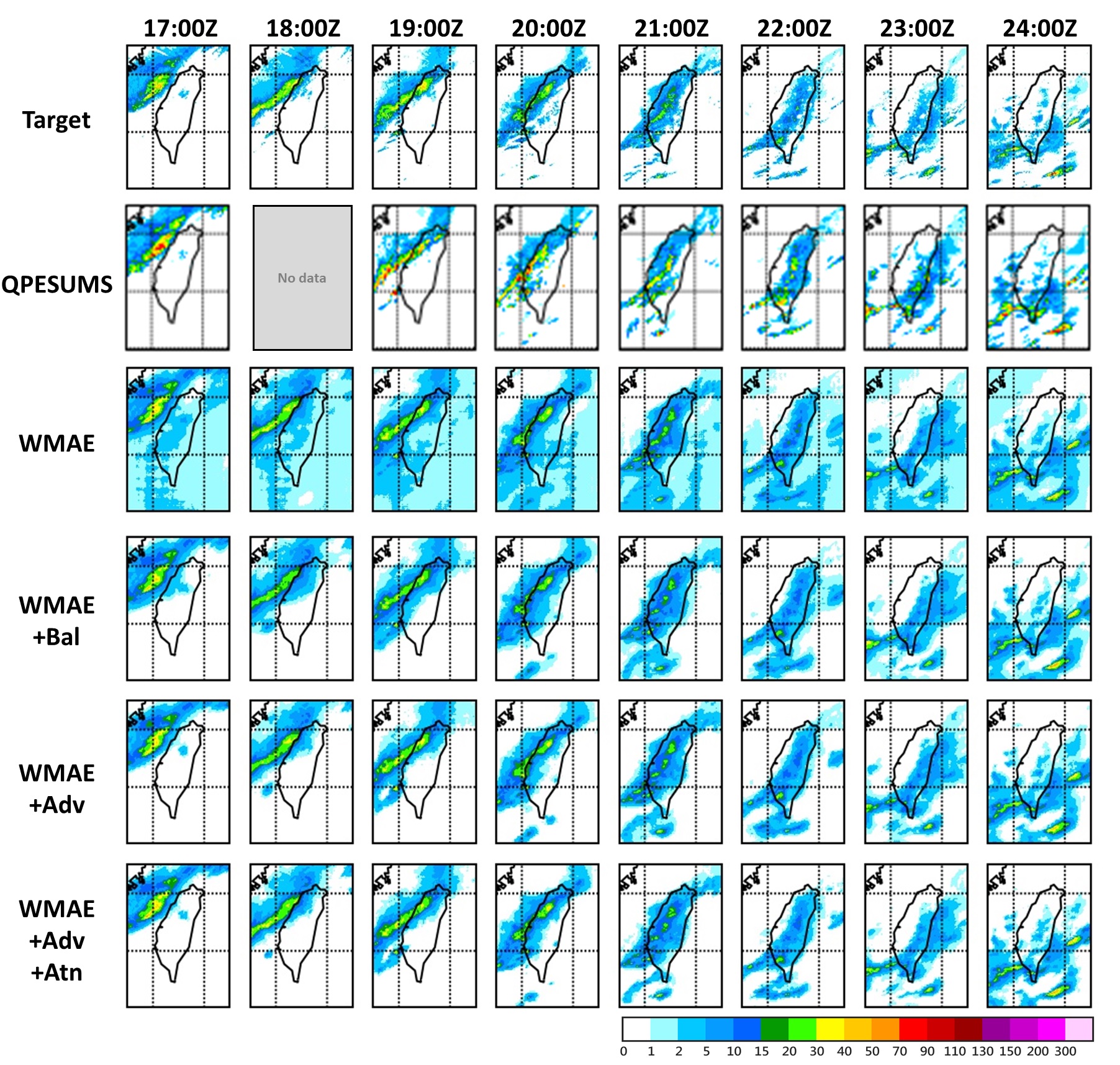}
  \caption{The hourly rainfall observation (first row) and predictions based on the benchmark operational QPESUMS radar echo extrapolation (second row) and four of our deep learning QPN models for the front event near Taiwan on $7_{th}$ May 2018 .}
  \label{fig:Fig_case}
\end{figure*}

A case study of the frontal system near Taiwan on 7th May 2018 illustrates the performance of QPN models developed in this study (Fig. ~\ref{fig:Fig_case}). 
The benchmark operational QPESUMS radar echo extrapolation (Fig. ~\ref{fig:Fig_case}, second row) fairly capture the motion of the frontal rainband but overpredict the maximum rainfall (over 70 mm), while the maximum rainfall in the observation (Fig. ~\ref{fig:Fig_case}, first row) is around 40 mm.
The GRU+WMAE model (Fig. ~\ref{fig:Fig_case}, third row) captures the rainband movement and the maximum rainfall but overforecast the light rain, leading to unrealistic large raining areas.
Furthermore, the GRU+WMAE+Bal model (Fig. ~\ref{fig:Fig_case}, fourth row) and the GRU+WMAE+Adv model (Fig. ~\ref{fig:Fig_case}, fifth row) handles the problem of predicting too large raining areas with GRU+WMAE+Adv doing a better job visually as well. After fixing this issue, the deep learning QPN model becomes competitive to the state-of-the-art operational model. 

By adding the attention mask into the model, the GRU+WMAE+Adv +Atn model performs better for the second and third-hour predictions. Specifically, the light rain area is better depicted for the GRU+WMAE +Adv+Atn model  (Fig. ~\ref{fig:Fig_case3h}), while other models have too much-blurred prediction in which the frontal system is less identifiable. Moreover the GRU+WMAE+Adv+Atn model retains the frontal rainband cells in the third-hour prediction as shown in (Fig. ~\ref{fig:Fig_case3h}, red ellipse). 

The performance diagram (Fig. ~\ref{fig:PD_1hr}) also suggests that, for this frontal case, our models outperform the operational QPESUMS extrapolation technique. For every rainfall thresholds except for 1 mm, our models have higher CSI scores (Fig. ~\ref{fig:PD_1hr}, green contours) than that of the QPESUMS.
For 1, 3, 5, and 10-mm thresholds, our modifications in the model (e.g., balanced loss, discriminator, and attention) gradually improve the CSI by increasing the success ratio (Fig. ~\ref{fig:PD_1hr}, X-axis) and keeping the probability of detection high. For higher 30, 40, and 50-mm thresholds, the GRU+WMAE+Adv+Atn model (Fig. ~\ref{fig:PD_1hr}, purple text) also outperforms other models.

\begin{table*}
  \begin{tabular}{c| cccccc| cccccc}
    \toprule
    Model & \multicolumn{6}{c}{2018}& \multicolumn{6}{c}{August 2018}\\
    & \multicolumn{3}{c}{WMAE(Th=0.5)} & \multicolumn{3}{c}{WMAE(Th=0.0)} &\multicolumn{3}{c}{WMAE(Th=0.5)} & \multicolumn{3}{c}{WMAE(Th=0.0)} \\
    & H0&H1&H2& H0&H1&H2& H0&H1&H2& H0&H1&H2\\
    \midrule
TrajGRU + WMSE+WMAE&0.015&0.011&0.015&0.3&0.708&0.71&0.08&0.04&0.08&0.2&0.57&0.61\\
TrajGRU + WMAE&0.009&0.002&0.003&0.252&0.372&0.397&0.05&0.01&0.03&0.12&0.01&0.13\\
GRU + WMAE& 0.005&0.005&0.006&0.102&0.152&0.159&0.03&0.03&0.02&0.07&0.08&0.06\\
GRU + WMAE+Bal&0.003&0.007&0.014&0.017&0.027&0.026&0.03&0.06&0.1&0.06&0.09&0.1\\
GRU + WMAE+Adv&0.007&0.007&0.012&0.017&0.061&0.119&0.04&0.05&0.09&0.06&0.12&0.21\\
GRU +WMAE+Atn&0.003&0.004&0.01&0.293&0.38&0.275&0.03&0.05&0.09&0.15&0.21&0.28\\
GRU + WMAE+Adv+Atn&0.002&0.004&0.007&0.011&0.045&0.093&0.02&0.02&0.03&0.04&0.11&0.15\\
    \bottomrule
  \end{tabular}
  \caption{Standard error on WMAE metric for first (H0), second (H1) and third (H2) hour prediction.}
   \label{tab:wmae_std}

\end{table*}

\section{Conclusion and Future Works}
In this work, we used last hour rain and radar data with a granularity of 10 minutes to predict 3 hourly rain rate maps, one for each hour. Deep learning approaches in general respond to uncertainty by predicting blurred predictions. We made use of a Discriminator to lessen redundant uncertainty thereby generating clearer rain-maps without adversely affecting the prediction quality for most rain-rate thresholds. After inspecting latest available rain map, it is relatively easy for a human to locate few regions where there is very less probability of rain. Deep learning models predict non-zero rain-rate on such regions as well which we refer to as redundant uncertainty. We further improved the performance at different thresholds by dynamically re-scaling the prediction using an attention module. We see the improvement at multiple rain-rate thresholds. We observed that on using real rain data as in our case, using a MSE based loss led to inferior performance and thus we easily outperformed~\cite{shi_deep_2017}. We still outperformed~\cite{shi_deep_2017} even after using it with an improved loss function (WMAE). Our model also performed better over QPESUMS, a benchmark used by meteorologists. Finally, we provide 3 hour, hourly rain-rates which is what is preferred by meteorologists over short duration predictions as done in many approaches including~\cite{shi_deep_2017}. Note that while multiple sequence predictions of such models could be averaged to get what we are predicting, we argue that directly optimizing hourly rain-rate would give our approach an edge for estimating hourly rain-rate.

There are a number of areas where we feel more work needs to be done.  One area of improvement could be to predict longer hours into the future. Another area of work can be to tackle uncertainty with multiple predictions. Instead of predicting one blurry rain-map, an architecture may be developed which predicts K probable clear rain-maps. Generally speaking, the field of precipitation nowcasting has immense scope of improvement from a machine learning perspective and we hope to be part of it.

\bibliographystyle{ACM-Reference-Format}
\bibliography{sample-base}

\appendix
\clearpage

\section{Supplemental}

\subsection{Utility of Threshold in WMAE}
We see the threshold being used in~\cite{shi_deep_2017} where they articulate it in their definition of mask. For establishing the utility of using a threshold, we trained  the model 'GRU+WMAE', where $WMMAE(Th=-1.0)$ was used for optimization. In this situation, as can be seen from Equation~\ref{eq:loss_weight}, the model is also penalized for its mis-predictions on non-rainy pixels as well. We see significant worse performance on $WMMAE(Th=0.5)$ which is our metric for evaluating heavy rainfall conditions in Table~\ref{tab:wmae_suppl} (first row). 

\subsection{Utility of Radar and Rain Data}
We trained 'GRU+WMAE+Adv' in two configurations where we varied the input data. In one, model was trained using just rain data. In other, model was trained with just radar data.  We found that model trained with just Radar data has considerably worse performance on WMAE metric as can be seen in Table~\ref{tab:wmae_suppl}. We observed model trained with just rain data to suffer from visual artefacts (very high intensity rain predictions on very small isolated regions) for one configuration($w_{adv}=0.05$).

\subsection{Blended Model}
In order to demonstrate that it is relatively easy to get good performance on low thresholds, we created a blended version of the model. We first create a classification model. It has the same Encoder-decoder structure as our Predictive module with sigmoid being the final activation. We train it using binary crossentropy loss. For creating pixelwise 0-1 labels, we use a threshold of 0.5mm/hr--- a pixel has label 1 if its hourly rain-rate is greater than the threshold and 0 otherwise. The classifier generates 3 probablity maps, one for each hour. For every consecutive hour, due to increase in uncertainty, the probability maps have consecutively lower value. So we rescale them and clip them in 0-1 region. If $W$ is the rescaled probablity map, $\hat{Y}_{Atn}$ and $\hat{Y}_{10min}$ are the predictions of our attention based model and last 10 minute rain respectively, the final prediction \(\hat{Y}=W*\hat{Y}_{Atn} + (1-W)*\hat{Y}_{10min}\). Note that we did not prefer the blending in the first place since besides being a patchy solution, it is also inferior on WMAE metric which is shown in Table~\ref{tab:wmae_suppl} (last column).
\begin{figure}[h]
  \centering
  \includegraphics[width=0.7\linewidth]{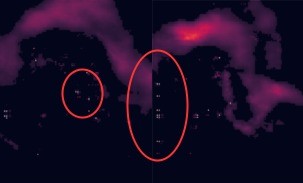}
  \caption{Artefacts observed in predicted rain maps in GRU+WMAE+Adv model trained on just rain data.   }
 \label{fig:artefacts}
\end{figure}
\begin{table}[h]
    \centering
    \begin{tabular}{cccc}
    \toprule
         Model&H0&H1&H2  \\
         \midrule
         GRU+WMAE(Th=0)& 0.49&0.90&1.08\\
         GRU+WMAE+Adv+ Just Radar &0.71&0.94&1.02\\
         Blended &2.66&4.92&5.70\\
          \bottomrule
    \end{tabular}
    \caption{Performance on WMAE(Th=0.5)}
    \label{tab:wmae_suppl}
\end{table}

\begin{table*}
    \centering
    \begin{tabular}{c |cccccccc| cccccccc}
    \toprule
    Model & \multicolumn{8}{c}{CSI} & \multicolumn{8}{c}{HSS}\\
    &1&3&5&10&15&20&30&40 &1&3&5&10&15&20&30&40 \\
    \midrule
    Last 10min         &0.27&0.25&0.23&0.19&0.17&0.16&0.14&0.11&   0.41&0.36&0.33&0.26&0.23&0.21&0.18&0.14\\
    Last 20min         &0.26&0.23&0.21&0.16&0.14&0.13&0.11&0.08&    0.39&0.32&0.29&0.22&0.19&0.17&0.14&0.11\\
    QPESUMS &0.19&0.16&0.13&0.12&0.11&0.1&0.08&0.07&   0.28&0.22&0.18&0.16&0.14&0.12&0.10&0.08\\
    TrajGRU + WMSE+WMAE&0.00&0.05&0.11&0.14&0.14&0.13&0.11&0.07&    0.04&0.07&0.15&0.20&0.19&0.18&0.15&0.10\\
    TrajGRU + WMAE     &0.01&0.14&0.19&0.19&0.17&0.16&0.13&0.10&    0.04&0.19&0.28&0.27&0.25&0.22&0.19&0.14\\
    GRU + WMAE         &0.01&0.17&0.19&0.19&0.17&0.16&0.14&0.10&    0.05&0.25&0.28&0.27&0.25&0.23&0.19&0.14\\
    GRU + WMAE+Bal&0.15&0.19&0.19&0.18&0.17&0.15&0.13&0.09&  0.22&0.28&0.28&0.27&0.24&0.22&0.18&0.13 \\
    GRU + WMAE+Adv&0.22&0.21&0.20&0.18&0.17&0.15&0.12&0.09&   0.33&0.31&0.28&0.27&0.24&0.21&0.17&0.13 \\
    GRU +WMAE+Atn&0.01&0.11&0.2&0.2&0.19&0.18&0.15&0.11&    0.04&0.16&0.29&0.3&0.27&0.25&0.21&0.15\\
    GRU + WMAE+Adv+Atn&0.22&0.22&0.21&0.20&0.19&0.17&0.15&0.11&  0.32&0.32&0.31&0.30&0.27&0.25&0.21&0.15 \\
    \bottomrule
  \end{tabular}
    \caption{HSS and CSI score for 2018 on first hour prediction}
    \label{tab:hss_csi_2018}
\end{table*}

\begin{table*}
    \centering
    \begin{tabular}{c |cccccc| cccccc}
    \toprule
    Model & \multicolumn{6}{c}{CSI} & \multicolumn{6}{c}{HSS}\\
    &1&5&10&15&20&30 &1&5&10&15&20&30 \\
    \midrule
    TrajGRU + WMSE+WMAE&    0&0.014&0.003&0.003&0.005&0.007&0&0.018&0.006&0.005&0.007&0.01\\
    TrajGRU + WMAE&         0.004&0.007&0.006&0.007&0.008&0.009&0.004&0.011&0.009&0.011&0.012&0.014\\
    GRU + WMAE&         0.003&0.002&0.005&0.006&0.007&0.009&0.003&0.003&0.007&0.009&0.011&0.013\\
    GRU + WMAE+Bal&         0.009&0.003&0.003&0.003&0.002&0.003&0.014&0.005&0.004&0.005&0.004&0.004\\
    GRU + WMAE+Adv& 0.012&0.004&0.006&0.007&0.007&0.007&0.019&0.007&0.008&0.011&0.01&0.01\\
    GRU +WMAE+Atn& 0.003&0.003&0.005&0.006&0.006&0.01&0.003&0.006&0.007&0.009&0.008&0.014\\
    GRU + WMAE+Adv+Atn&0.008&0.003&0.004&0.007&0.008&0.007&0.013&0.006&0.006&0.01&0.011&0.01\\

    \bottomrule
  \end{tabular}
    \caption{Standard Error of HSS and CSI score for 2018 on first hour prediction}
    \label{tab:hss_csi_2018_std}
\end{table*}

\begin{table*}
    \centering
    \begin{tabular}{c |cccccccc| cccccccc}
    \toprule
    Model & \multicolumn{8}{c}{CSI} & \multicolumn{8}{c}{HSS}\\
    &1&3&5&10&15&20&30&40 &1&3&5&10&15&20&30&40 \\
    \midrule
    Last 20min&        0.35&0.33&0.31&0.27&0.23&0.2&0.15&0.12&    0.6&0.53&0.49&0.4&0.32&0.26&0.19&0.15 \\
    Last 10min&        0.36&0.34&0.33&0.29&0.25&0.22&0.17&0.14&   0.62&0.57&0.52&0.43&0.36&0.3&0.22&0.18 \\
    QPESUMS& 0.31&0.3&0.28&0.24&0.2&0.17&0.12&0.09&0.55&0.48&0.44&0.35&0.27&0.22&0.15&0.11 \\
    TrajGRU + WMSE+WMAE& 0.00&0.16&0.27&0.29&0.28&0.25&0.18&0.12&   0.18&0.28&0.42&0.45&0.41&0.35&0.25&0.17\\
    TrajGRU + WMAE&      0.05&0.31&0.36&0.35&0.31&0.28&0.21&0.16& 0.21&0.51&0.59&0.56&0.48&0.41&0.3&0.22\\
    GRU + WMAE          &0.06&0.33&0.36&0.34&0.31&0.28&0.21&0.15& 0.23&0.56&0.59&0.55&0.48&0.41&0.3&0.22\\
    GRU + WMAE+Bal&    0.25&0.34&0.36&0.34&0.31&0.28&0.21&0.15&   0.44&0.57&0.59&0.55&0.47&0.41&0.29&0.21\\
    GRU + WMAE+Adv&    0.34&0.36&0.36&0.34&0.31&0.27&0.2&0.14&  0.59&0.6&0.6&0.54&0.47&0.4&0.28&0.2\\
    GRU +WMA+Atn&0.04&0.31&0.36&0.35&0.32&0.29&0.22&0.16&  0.21&0.51&0.59&0.56&0.49&0.43&0.32&0.23\\
    GRU + WMAE+Adv+Atn&0.32&0.36&0.36&0.35&0.32&0.29&0.23&0.16& 0.56&0.6&0.61&0.57&0.5&0.43&0.32&0.23\\
    \bottomrule
  \end{tabular}
    \caption{HSS and CSI score for Highest Rainfall Month in 2018, namely August  on first hour prediction}
    \label{tab:hss_csi_2018Aug}
\end{table*}
\begin{table*}
    \centering
    \begin{tabular}{c |cccccc| cccccc}
    \toprule
    Model & \multicolumn{6}{c}{CSI} & \multicolumn{6}{c}{HSS}\\
    &1&5&10&15&20&30 &1&5&10&15&20&30 \\
    \midrule
TrajGRU + WMSE+WMAE& 0.002&0.011&0.009&0.006&0.007&0.01&0.001&0.018&0.017&0.012&0.012&0.015\\
TrajGRU + WMAE&  0.012&0.004&0.003&0.005&0.007&0.01&0.011&0.008&0.006&0.01&0.014&0.016\\
GRU + WMAE&   0.013&0.001&0.002&0.004&0.006&0.008&0.012&0.001&0.004&0.008&0.011&0.013\\
GRU + WMAE+Bal& 0.017&0.003&0.001&0.003&0.003&0.004&0.024&0.007&0.003&0.005&0.006&0.006\\
GRU + WMAE+Adv& 0.012&0.003&0.003&0.003&0.005&0.004&0.02&0.007&0.006&0.006&0.008&0.007\\
GRU +WMA+Atn& 0.014&0.004&0.001&0.001&0.003&0.008&0.012&0.009&0.003&0.002&0.005&0.011\\
GRU + WMAE+Adv+Atn& 0.009&0.002&0.001&0.003&0.004&0.006&0.016&0.005&0.002&0.007&0.008&0.01\\
    \bottomrule
  \end{tabular}
    \caption{Standard Error of HSS and CSI score for August 2018 on first hour prediction}
    \label{tab:hss_csi_2018Aug_std}
\end{table*}

\end{document}